\documentclass[11pt]{article}

\usepackage[preprint]{acl}
\usepackage{CJKutf8}

\usepackage{times}
\usepackage{latexsym}
\usepackage{amsmath}
\DeclareMathOperator*{\argmax}{arg\,max}
\DeclareMathOperator*{\argmin}{arg\,min}

\usepackage[T1]{fontenc}

\usepackage[utf8]{inputenc}

\usepackage{microtype}

\usepackage{inconsolata}

\usepackage{graphicx}

\newcommand{\ours}{PIAST}
\newtheorem{remark}{Remark}

\usepackage{hyperref}
\usepackage{url}
\usepackage{booktabs}        
\usepackage{graphicx}         
\usepackage[table,dvipsnames]{xcolor}
\usepackage{array}         
\usepackage{standalone} 
\usepackage{textcomp}         
\usepackage[font=small,labelfont=bf]{caption}
\usepackage{wrapfig}

\usepackage{enumitem}
\setlist[itemize]{leftmargin=7pt, itemsep=0pt, topsep=0pt}

\usepackage{xcolor}
\usepackage[most]{tcolorbox}
\usepackage{tikz}
\usetikzlibrary{calc}

\definecolor{royal}{HTML}{3A0CA3}
\definecolor{iris}{HTML}{5B3BE0}
\definecolor{fuchsia}{HTML}{D946EF}
\definecolor{teal}{HTML}{14B8A6}
\definecolor{sky}{HTML}{0EA5E9}
\definecolor{emerald}{HTML}{10B981}
\definecolor{amber}{HTML}{F59E0B}
\definecolor{coral}{HTML}{FF6B6B}
\definecolor{rose}{HTML}{F43F5E}
\definecolor{lavender}{HTML}{7C4DFF}
\definecolor{slate}{HTML}{334155}
\definecolor{cocoa}{HTML}{7B3F00}

\newcommand{\promptbox}[3]{%
  \begin{tcolorbox}[
    enhanced, breakable,
    colback=white,
    frame hidden, boxrule=0pt,
    left=6pt, right=6pt, top=6pt, bottom=6pt,   
    before skip=4pt, after skip=4pt,            
    arc=1mm,                                    
    borderline west={2pt}{0pt}{#3!85!black},    
    title={\strut #1},
    fonttitle=\bfseries\sffamily\footnotesize,  
    coltitle=white,
    attach boxed title to top left={yshift=-1.2mm, xshift=6pt},
    boxed title style={
      boxrule=0pt,
      colback=#3!85!black,                      
      left=5pt, right=5pt, top=2pt, bottom=2pt, 
      arc=0.8mm
    },
    overlay={
      \draw[line width=0.3pt, draw=#3!35!black, rounded corners=0.6mm]
        ($(interior.north west)+(0.35mm,-0.35mm)$)
        rectangle
        ($(interior.south east)+(-0.35mm,0.35mm)$);
    }
  ]
  {\footnotesize #2} 
  \end{tcolorbox}%
}

\usepackage{adjustbox}
\newcolumntype{:}{!{\vrule width 0.8pt}}

\usepackage{algorithm}
\usepackage{algpseudocode}
\usepackage{amsmath, amssymb}
\newcommand{\meanstd}[2]{#1\textsubscript{\textcolor{gray}{\,\(\pm\)\,#2}}}

\newcommand{\AvgCell}[1]{\csname avg@#1@cell\endcsname}

\usepackage{booktabs,array,colortbl,xcolor,makecell,tikz}
\usetikzlibrary{decorations.pathmorphing}

\definecolor{headerbg}{HTML}{0EA5A2} 
\definecolor{headerfg}{HTML}{FFFFFF} 
\definecolor{rowalt}{HTML}{F8FAFC}   
\definecolor{yescol}{HTML}{16A34A}   
\definecolor{nocol}{HTML}{DC2626}    
\definecolor{maybe}{HTML}{F59E0B}    
\definecolor{pillbg}{HTML}{E5E7EB}   

\renewcommand{\arraystretch}{1.25}
\newcommand{\Badge}[1]{%
  \tikz[baseline=-0.6ex, scale=0.22]{
    \draw[line width=0.9pt, draw=black!12, fill=black!2] (0,0) circle (1.9);
    #1
  }%
}

\newcommand{\YesMark}{%
\scalebox{0.5}{
  \Badge{%
    \draw[line width=2.2pt, draw=yescol, line cap=round, line join=round]
      (-1.0,-0.2) -- (-0.2,-1.0) -- (1.1,0.9);
  }%
  }
}

\newcommand{\NoMark}{%
\scalebox{0.5}{
  \Badge{%
    \draw[line width=2.0pt, draw=nocol, line cap=round] (-1.1,-1.1) -- (1.1,1.1);
    \draw[line width=2.0pt, draw=nocol, line cap=round] (-1.1,1.1) -- (1.1,-1.1);
  }%
  }
}

\newcommand{\WaveMark}{%
\scalebox{0.5}{
  \Badge{%
    \draw[
      line width=2.0pt, draw=maybe,
      decorate, decoration={snake, amplitude=0.55pt, segment length=3.2pt}
    ] (-1.3,0) -- (1.3,0);
  }%
  }
}

\newcommand{\SpeedBar}[1]{%
  \tikz[baseline=-0.6ex, x=1.1ex, y=1.0ex]{
    \foreach \i in {1,...,5}{
      \path (\i,0) node[draw=black!20, fill=pillbg, rounded corners=0.9pt, minimum width=1.0ex, minimum height=1.15ex, inner sep=0pt] (s\i) {};
    }
    \foreach \i in {1,...,#1}{
      \fill[yescol] (s\i.south west) rectangle (s\i.north east);
      \draw[rounded corners=0.9pt, line width=0.35pt, draw=yescol!70!black] (s\i.south west) rectangle (s\i.north east);
    }
  }%
}

\newcommand{\SpeedExtSlow}{\SpeedBar{1}}
\newcommand{\SpeedLow}{\SpeedBar{2}}
\newcommand{\SpeedMed}{\SpeedBar{3}}
\newcommand{\SpeedVeryFast}{\SpeedBar{4}}
\newcommand{\SpeedExtFast}{\SpeedBar{5}}

\newcolumntype{C}{>{\centering\arraybackslash}m{2.5cm}}

\usetikzlibrary{positioning,arrows.meta}

\def\redc{\cellcolor[HTML]{FF999A}}
\def\orangec{\cellcolor[HTML]{FFCC99}}
\def\yellowc{\cellcolor[HTML]{FFF8AD}}
\usepackage[table]{xcolor}
\definecolor{lightyellow}{RGB}{255, 255, 224}

\title{PIAST: Rapid \underline{P}rompting with \underline{I}n-context \underline{A}ugmentation for \underline{S}carce \underline{T}raining data}

\author{
\begin{tabular}{@{}c@{\hspace{2.5em}}c@{}}
Pawe{\l} Batorski & Paul Swoboda \\
\multicolumn{2}{c}{{\normalfont Heinrich Heine Universit\"at D\"usseldorf}} \\
\multicolumn{2}{c}{{\normalfont\texttt{\{pawel.batorski,paul.swoboda\}@hhu.de}}}
\end{tabular}
}

\begin{document}
\maketitle

\begin{figure*}[ht]
  \centering
  \includegraphics[width=\linewidth]{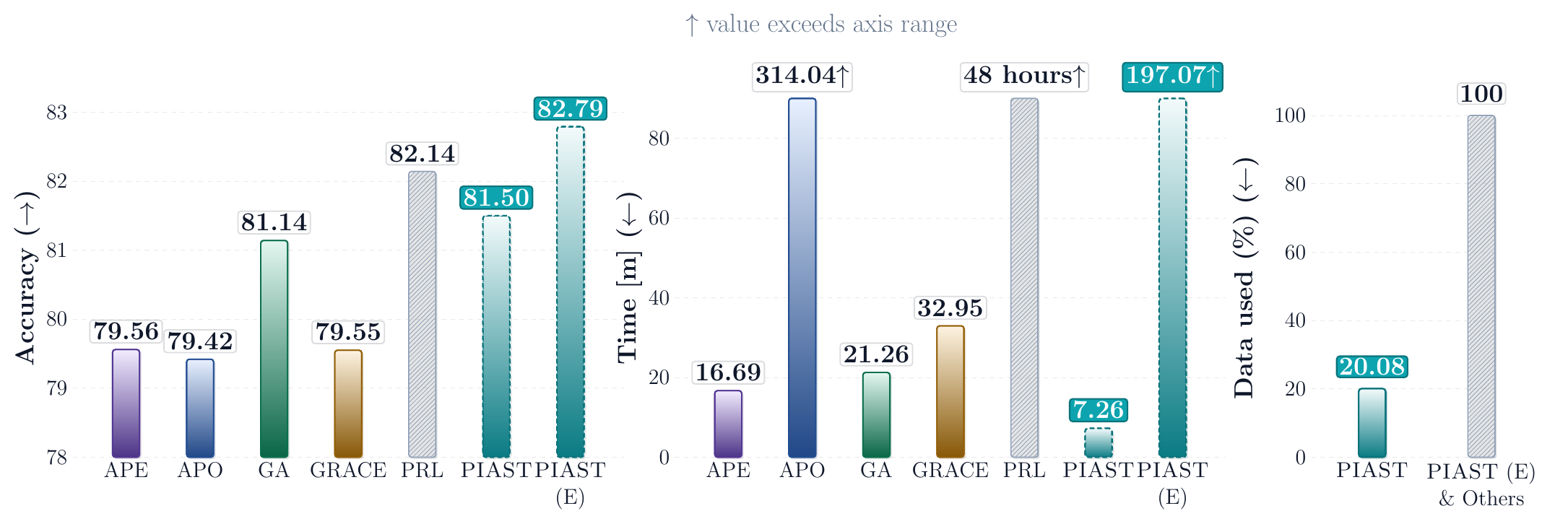}
\caption{Overview of the results averaged over seven different text classification tasks, each run three times, comparing \ours{} against current benchmarks. \ours{} is able to generate high-quality prompts very efficiently, while requiring only a small portion of the dataset yielding comparable results to the current SOTA methods.}
 \label{fig:teaser}
\end{figure*}

\begin{abstract}
LLMs are highly sensitive to prompt design, but handcrafting effective prompts is difficult and often requires intricate crafting of few-shot examples. 
We propose a fast automatic prompt construction algorithm that augments human instructions by generating a small set of few shot examples.
Our method iteratively replaces/drops/keeps few-shot examples using Monte Carlo Shapley estimation of example utility.
For faster execution, we use aggressive subsampling and a replay buffer for faster evaluations.
Our method can be run using different compute time budgets. 
Under a limited budget, it outperforms prior automatic prompting methods on text simplification and mathematical reasoning (GSM8K, DeepMath, Math500), while achieving second-best results on classification and summarization and third-best on MedQA.
With an extended, yet still modest budget, \ours{} sets a new state of the art among automatic prompting methods on classification, simplification, GSM8K, DeepMath, and Math500.
Overall, our results suggest that optimizing in-context examples, rather than exhaustively searching over instruction rewrites is the dominant lever for fast and data-efficient prompt engineering.
We will release code and data upon acceptance.
\end{abstract}

\section{Introduction}
\label{sec:intro}
Automatic prompt engineering has emerged as a practical way to adapt LLMs without gradient updates.
However, many existing methods are impractical in time and data constrained settings: (i)~some require hours of compute to explore a large prompt search space, and (ii)~they rely on sizeable training sets to reliably score candidates. 

\noindent Recent work on prompt generation for previously unseen tasks~\citep{batorski2025gps} alleviates the per-task tuning burden, but still lags behind methods that are optimized separately for each task. Moreover, most prior work optimizes only the instruction string, for example via rephrasing (using e.g.\ evolutionary algorithms or extensive search), ignoring the most impactful component of in-context learning (ICL): few-shot examples.
The main exceptions are~\citep{batorski2025prl}, which synthesizes examples, but its computational cost is dozens of hours, and~\citep{pryzant2023automatic}, which only selects examples from the training set.
\noindent 
Our method is, to our knowledge, the first method that is fast, synthesizes new few-shot examples not found in the training set and requires relatively less access to training examples.
When run long enough, our method additionally obtains new state of the art results among automatic prompting methods for a number of tasks. 
\noindent 
Our method works as follows:
We first synthesize a proposal set of in-context examples that we append to our initial prompt.
Then, in our optimization loop, we evaluate its efficacy on a small randomized evaluation set and identify the least helpful examples using a Monte Carlo Shapley estimator and replace, drop or keep it.
If replaced, we draw from a pool of newly proposed few-shot examples.
For efficiency and stability, we use a replay buffer for the evaluation set.
\noindent 
Our algorithm has a favorable anytime performance: 
When run with a small computational budget, we attain second best results among our baselines on classification and summarization and already exceeds previous SoTA on simplification and mathematical reasoning tasks.
When run with an extended budget that is still comparable to some other baselines, we exceed previous methods additionally on classification.
Interestingly, even when running without the iterative update loop and only using the first generated few-shot examples, we often still get competitive results.
\noindent 
To summarize, our contributions are as follows:
\begin{description}[leftmargin=!, labelwidth=10pt]
\item[Conceptual:] We propose \ours{}, an automatic prompt construction method that augments a concise human-written instruction with a small set of automatically generated few-shot examples.
We use an iterative improvement loop that improves the current set of few-shot examples using Shapley values to estimate utility of individual examples.
\item[Implementation:] For a fast implementation, we approximate Shapley values, KV-cache reuse for shared ICL prefixes, PagedAttention for compact KV memory management and continuous token-level batching to maintain high GPU utilization.
\item[Empirical Results:] We demonstrate strong performances using the same set of robust hyperparameters on text classification, summarization and simplification as well as GSM8K.
Our approach yields strong anytime performance: When using only a subset of data and a small computational budget we obtain SoTA on text simplification and GSM8K and obtain second best results on summarization and classification. With an extended budget and full training set access we additionally set a new SoTA on classification among automatic prompting methods.
\end{description}

\begin{table*}[h!]
\centering
\begingroup
\rowcolors{2}{rowalt}{white}

\begin{adjustbox}{max width=0.97\textwidth}
\begin{tabular}{@{} l C C C C c @{}}
\rowcolor{headerbg}
\multicolumn{1}{@{}l}{\color{headerfg}\bfseries Algorithm} &
\multicolumn{1}{c}{\color{headerfg}\bfseries Dataset} &
\multicolumn{1}{c}{\color{headerfg}\bfseries Refinement} &
\multicolumn{1}{c}{\color{headerfg}\bfseries Few\mbox{-}shot} &
\multicolumn{1}{c}{\color{headerfg}\bfseries Auto Gen.} &
\multicolumn{1}{c@{}}{\color{headerfg}\bfseries Speed} \\
\midrule
Manual Instruction~\citep{zhang2022opt} & \NoMark  & \NoMark   & \NoMark   & \NoMark  & \SpeedExtFast \\
APE~\citep{zhou2022large}                & \YesMark & \NoMark   & \NoMark   & \YesMark & \SpeedMed     \\
APO~\citep{pryzant2023automatic}                & \YesMark & \YesMark  & \WaveMark & \YesMark & \SpeedLow     \\
EvoPrompt~\citep{guo12connecting}          & \YesMark & \YesMark  & \NoMark   & \YesMark & \SpeedMed    \\
PRL~\citep{batorski2025prl}                & \YesMark & \YesMark  & \YesMark  & \YesMark & \SpeedExtSlow \\
GRACE~\citep{grace}                & \YesMark & \YesMark  & \NoMark  & \YesMark & \SpeedMed \\
\ours{}            & \WaveMark& \YesMark  & \YesMark  & \YesMark & \SpeedVeryFast\\
\bottomrule
\end{tabular}
\end{adjustbox}
\endgroup

\vspace{4pt}
{\footnotesize
\textbf{Legend:}\;
\YesMark\ yes \;\;
\NoMark\ no \;\;
\WaveMark\ partial \;\;
Speed:\; \SpeedExtSlow\ slow, $\dots$, \SpeedExtFast\ fast.
}
\caption{Comparison of \ours{} with other methods from the literature. 
Dataset indicates the fraction of the dataset used during construction of the prompt. 
Refinement shows whether the method iteratively improves the current prompt or generates a new one in a single step. 
Few-shot specifies whether the method is capable of generating few-shot examples.
Auto Gen. denotes whether prompts are generated automatically.}
\label{tab:comparison}
\end{table*}

\section{Related Work}
\textbf{Prompt Engineering}
improves model capabilities without retraining~\citep{liu2023pretrain}. Chain-of-Thought (CoT)~\citep{wei2022chain} elicits intermediate reasoning; Tree-of-Thought (ToT)~\citep{yao2023tree} explores multiple paths; Program-of-Thoughts~\citep{chen2022program} and Graph-of-Thoughts~\citep{besta2024graph} structure prompts as programs/graphs. Least-to-Most prompting decomposes problems into subproblems~\citep{zhou2023leasttomost}, and zero-shot CoT and self-consistency improve robustness~\citep{kojima2022large,wang2022self}. %
Few-shot prompting~\citep{brown2020language} conditions on in-prompt exemplars and is effective for puzzles and evidence extraction~\citep{xu2023llms,greenblatt2024getting,sivarajkumar2024empirical}. Lu et al.~\citep{lu2022fantastically} show strong sensitivity to demonstration order and propose more robust orderings.
\noindent
\paragraph{Automated Prompt Engineering} replaces manual prompt design with automated search and refinement. %
APE~\citep{zhou2022large} generates candidate prompts and selects best by performance, remaining purely generative. %
APO~\citep{pryzant2023automatic} iteratively refines prompts via natural-language critiques, but its few-shot examples are restricted to the training set. %
EvoPrompt~\citep{guo12connecting} evolves prompts with evolutionary operators, while PromptAgent~\citep{wang2023promptagent} frames prompt optimization as planning and applies MCTS with error-driven feedback. %
Promptbreeder~\citep{fernando2023promptbreeder} studies self-referential prompt evolution, where prompts generate and select improved variants of themselves via an evolutionary loop. 
Similarly, \citep{wang2025evolving} frame prompt optimization as an open-ended, self-replicating process. 
GRACE~\citep{grace} performs iterative prompt refinement with a gating rule and adaptively compresses prompts. 
OPRO~\citep{yang2024llm_optimizers} uses an LLM as a black-box optimizer over instructions given past candidates and scores. AutoPrompt~\citep{shin2020autoprompt} searches over discrete prompt tokens using gradient signals, and~\citep{lu2024strings} argue that random sampling is a strong prompt-optimization baseline.
\noindent
Other approaches leverage reinforcement learning, such as RLPrompt~\citep{deng2022rlprompt} (short token prompts) and PRL~\citep{batorski2025prl}, which can synthesize in-context examples when beneficial. Among these, only APO and PRL explicitly incorporate examples: APO reuses training examples, while PRL can introduce novel examples but often requires tens of hours of compute, limiting practicality. In contrast, \ours{} rapidly generates few-shot examples not present in the training data. We summarize comparisons against the baselines we test against in Table~\ref{tab:comparison}.

\paragraph{Multi-agent / multi-module prompt optimization.}
Recent work optimizes prompts for multi-stage LM programs and agentic pipelines rather than a single LM call. DSPy~\citep{dspy} provides a framework and ``teleprompting'' optimizers; MiPRO~\citep{mipro} and GePA~\citep{gepa} optimize instructions and demonstrations across modules; BetterTogether~\citep{bettertogether} combines prompt optimization with parameter updates in modular pipelines; and SGLang~\citep{sglang} targets efficient execution of structured LM programs. We do not compare to these methods because they assume a multi-module / multi-agent program and optimize end-to-end behavior, whereas \ours{} produces a single reusable prompt for one LLM call.

\paragraph{Demonstration selection / retrieval.}
Demonstration selection constructs a prompt per test input by retrieving demonstrations from a fixed pool, typically the labeled training set. \citep{rubin-etal-2022-learning} train a dense retriever using LM-scored (input, demo) helpfulness labels; \citep{li-etal-2023-unified} learn a retriever that generalizes across task families with a unified list-wise ranking objective; and Skill-KNN~\citep{an-etal-2023-skill} uses engineered representations for instance-wise selection. MoD~\citep{wang2024mixture} partitions the demo pool into expert groups for collaborative retrieval. We do not compare to these approaches because they require per-instance retrieval and continued access to a demo bank at inference time, whereas \ours{} generates a compact set of demonstrations once per task and reuses the same prompt for all inputs.

\section{Method}
\label{sec:method}

In this section, we present our method, which is composed of three components: the Example Proposer, the Prompt Evaluator, and the Example Improver. Each component is instantiated as a frozen LLM with a distinct role in the overall pipeline. Our final prompt consists of the hand-crafted instruction proposed by \citep{zhang2022opt}, concatenated with the in-context examples produced by our optimization procedure.

\paragraph{Example Proposer.}  
The Example Proposer is responsible for generating initial candidate examples. It receives a task-specific initial instruction and produces a set of examples accordingly. To ensure coverage and robustness, the generated examples are deliberately diverse in both topic and length. Each example is then subject to a subsequent replace/drop/keep decision. The prompt for the Example Proposer is given in Appendix \ref{sec:prompt_generator}.

\begin{figure}[ht]
  \centering
  \includegraphics[width=\columnwidth]{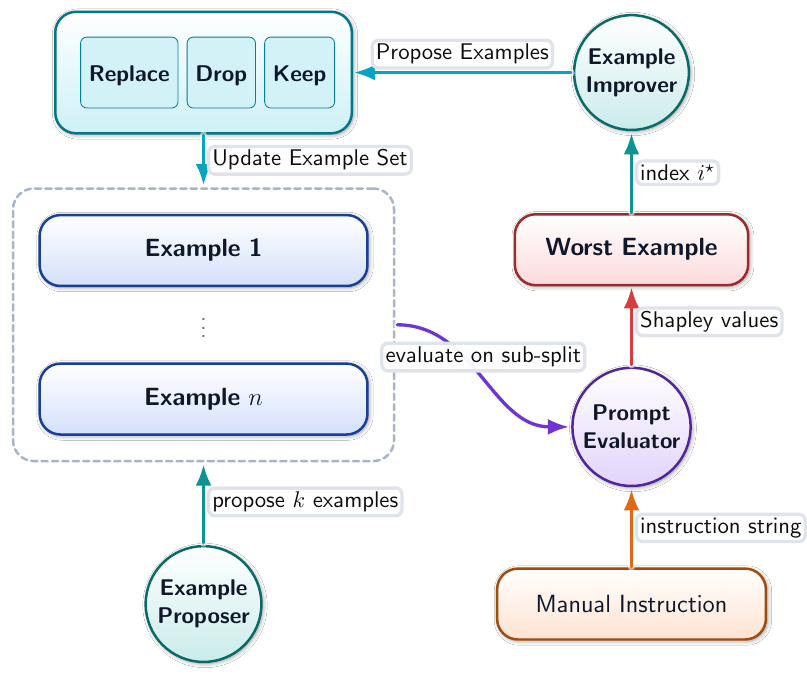}
\caption{
Pipeline of \ours{}.
Initially, the Example Proposer generates examples, which are then iteratively improved by evaluating them with the Prompt Evaluator and choosing new examples from the Example Improver to incorporate into the set of current in-context examples.
}
  \label{fig:method}
\end{figure}

\paragraph{Prompt Evaluator.}  
The Prompt Evaluator assesses the quality of candidate prompts. Given a prompt, it evaluates its performance on a subset $D$ of the data using a specified metric $f$. This step ensures that only the most effective prompts are selected for further use. In our setup, the evaluated prompt is the base instruction concatenated with the
proposed examples.

\paragraph{Example Improver.}  
The Example Improver starts from a current prompt and iteratively changes the in-context examples in a replace/drop/keep cycle. The examples produced may differ in structure, topic, and length, thereby increasing diversity in the candidate pool. This process mirrors the behavior of the Example Proposer, where randomness in topic and sentence length is also introduced to encourage exploration. The prompt for the Example Improver is given in Appendix \ref{sec:prompt_improver}.
\noindent 
The example improver uses Shapley values to estimate quality of examples and implements a replace/drop/keep cycle as detailed below.

\paragraph{Example Selection via Shapley Values.}  
At each iteration of the example improver, we identify the least useful example, which is then considered for a replace/drop/keep decision. To determine which, we employ Shapley values, which provide a principled way of estimating the contribution of each example to overall performance.

\noindent 
Let $[n]=\{1,\dots,n\}$ index the current few-shot examples and let
$v:2^{[n]}\!\to\mathbb{R}$ be a function that maps any subset
 $S\subseteq[n]$ to its utility (computed on $D$). In our setup, $v(\varnothing)$ is the accuracy of the instruction-only baseline (no few-shot examples). Write $\Pi([n])$ for the set of all permutations of $[n]$.
 For $\pi\in\Pi([n])$ and $i\in[n]$, define the predecessor set
 \[
 P_i^\pi \;=\; \{\, j\in[n] : j \text{ precedes } i \text{ in } \pi \,\}.
 \]

 The Shapley value of example $i$ is
 \begin{equation}
 \phi_i
 \;=\;
 \frac{1}{n!}\sum_{\pi\in\Pi([n])}
 \Bigl( v\bigl(P_i^\pi \cup \{i\}\bigr) - v\bigl(P_i^\pi\bigr) \Bigr).
 \label{eq:shapley}
 \end{equation}

\begin{remark}
Although in-context learning can be sensitive to the order of examples, our Shapley estimator is order-agnostic: it treats coalitions as sets and evaluates each subset using the current fixed example order used by the optimizer and at inference time. This is still meaningful because our algorithm never searches over orderings, only replaces or drops examples, so we want attributions under the deployed order. Averaging over random subsets still captures redundancy and complementarity between examples, giving a robust signal at low computational cost.
\end{remark}


\paragraph{Monte Carlo Approximation}
Because summing over all $n!$ permutations in \eqref{eq:shapley} is infeasible,
we approximate it by sampling $K$ i.i.d. permutations
$\pi^{(1)},\ldots,\pi^{(K)} \sim \mathrm{Unif}(\Pi([n]))$ and computing
\begin{equation}
 \widehat{\phi}_i
 \;=\;
 \frac{1}{K}\sum_{k=1}^{K}
 \Bigl( v\bigl(P_i^{\pi^{(k)}} \cup \{i\}\bigr) - v\bigl(P_i^{\pi^{(k)}}\bigr) \Bigr).
 \label{eq:mc-shapley}
 \end{equation}
 In practice, each permutation yields marginal contributions for all $i$ in a single sweep
 by maintaining the running predecessor set; averaging over $K$ permutations gives an
 unbiased estimator of $\phi_i$

\noindent 
The pseudo-code for selecting the least useful example is included in Appendix \ref{sec:pseudocode}.


\paragraph{Replace/Drop/Keep decision.}
Given the current set of in\mbox{-}context examples $[n]$ we determine the least helpful example $i^\star$ using the Shapley criterion:
\begin{equation}
i^\star \;=\; \argmin_{i\in [n]}\widehat{\phi}_i.
\label{eq:argmin}
\end{equation}
For this step, the Example Improver proposes $m$ candidate examples  $C = \{c_1, \ldots, c_m\}$ for potential appending to the current few shot example set.
To decide whether to replace, keep,  or drop the index $i^\star$, we compute the following scores:

\begin{align}
r &= \max_{c \in \mathcal{C}} v([n] \backslash \{i^\star\} \cup \{c\}) \tag{\textsc{replace}} \\
d &= v([n] \backslash \{i^\star\}) \tag{\textsc{drop}} \\
k &= v([n]) \tag{\textsc{keep}}
\end{align}
We select replace, keep or drop by checking whichever score is largest. 
When having ties, we prefer replace over drop over keep.
The next prompt becomes $(N\setminus\{i^\star\})\cup\{c^\star\}$ under \textsc{replace}, $N\setminus\{i^\star\}$ under \textsc{drop}, and $N$ under \textsc{keep}. This policy ensures we only adopt a modification when it does not underperform the best available alternative (drop or status quo).

\begin{remark}
Note that the Replace/Drop/Keep step could also be formulated directly using Shapley values. However, this would significantly increase the computational cost of each iteration, whereas our design prioritizes speed and efficiency.
\end{remark}

\paragraph{Replay Buffer}
Our method relies on sampling a subset of the training data at each iteration.
Consequently, newly crafted examples can overfit the current subset and fail to generalize to training subsets drawn in later iterations, since there is no mechanism enforcing that they also perform well on previously seen data. 
To mitigate this, after each iteration we store a small portion of the training data in a replay buffer. 
At the next iteration, this buffer is merged with the freshly sampled subset, which preserves accuracy across iterations by acting as a regularizer: newly crafted examples must also succeed on data sampled in prior iterations.

\paragraph{Speed.}
To make our implementation fast, we employ the following techniques:
We use a KV cache \citep{radford2019gpt2} to avoid recomputing attention over already-processed tokens: 
Keys and values for the shared in-context prefix are cached once and then reused across 
(i)~all tokens within a sequence and (ii)~multiple evaluation queries that share this prefix. 
In addition, we rely on PagedAttention \citep{korthikanti2023vllm} to store the KV cache in paged memory chunks, 
which minimizes fragmentation and data movement while enabling efficient, contiguous access during decoding. 
Finally, we leverage continuous (token-level) batching \citep{yu2022orca}, in which the scheduler dynamically 
forms a new batch at each decoding step by admitting fresh requests and retiring completed ones, thereby overlapping 
prefill and decoding and maintaining high GPU utilization.
\noindent 
Pseudocode for \ours{} can be found in Appendix \ref{sec:pseudocode}.

\section{Experiments}
\label{sec:exp}

All experiments are run on a single NVIDIA A100 GPU.
Unless stated otherwise, \ours{} uses Qwen2.5-7B-Instruct ~\citep{yang2024qwen2} for all three roles in the pipeline: the Example Proposer, the Example Improver, and the Prompt Evaluator.
For a controlled comparison, we evaluate all baselines using the same underlying LLM, so differences in results reflect the prompting algorithms rather than model choice.
\noindent 
We compare \ours{} against established automatic prompting methods across four task families:
(i)~text classification, (ii)~summarization, (iii)~mathematical reasoning, and (iv)~domain-specific question answering.
We additionally report results on text simplification in Appendix~\ref{sec:sim}.
\noindent 
We also conduct ablation studies to isolate the effect of key design and hyperparameter choices.
In each ablation, we vary a single factor while holding all others fixed.
All reported numbers (main results and ablations) are averaged over three independent runs.
Across all tasks, we use one shared hyperparameter configuration; the full setting is provided in Appendix~\ref{sec:hyperparameters}.

\begin{table*}[ht]
\tiny
\centering
\renewcommand{\arraystretch}{1.2}
\resizebox{1.0\textwidth}{!}{%
\begin{tabular}{cccccccc:c}
\hline
\textbf{Method / Dataset} & \textbf{SST-2} & \textbf{CR} & \textbf{MR} & \textbf{SST-5} & \textbf{AG's News} & \textbf{TREC} & \textbf{Subj} & \textbf{Avg} \\ 
\hline
MI
& 92.70 & 87.25 & 87.40 & 52.31 & 82.29 & 69.20 & 57.95 & 75.59 \\

NI
& \yellowc{95.77} & 91.50 & \yellowc{90.85} & 51.90 & 83.43 & 66.60 & 68.10 & 78.31 \\

APO
& 93.71\textsubscript{\textcolor{gray}{±0.25}}
& \redc{93.48\textsubscript{\textcolor{gray}{±0.24}}}
& 89.97\textsubscript{\textcolor{gray}{±1.37}}
& \yellowc{53.94\textsubscript{\textcolor{gray}{±0.29}}}
& 83.73\textsubscript{\textcolor{gray}{±0.31}}
& 71.30\textsubscript{\textcolor{gray}{±1.90}}
& 69.80\textsubscript{\textcolor{gray}{±5.96}}
& 79.42 \\

APE
& 91.23\textsubscript{\textcolor{gray}{±0.66}}
& \yellowc{92.87\textsubscript{\textcolor{gray}{±0.02}}}
& 89.90\textsubscript{\textcolor{gray}{±0.94}}
& 49.37\textsubscript{\textcolor{gray}{±5.66}}
& 82.58\textsubscript{\textcolor{gray}{±1.20}}
& \yellowc{77.07\textsubscript{\textcolor{gray}{±1.61}}}
& 73.92\textsubscript{\textcolor{gray}{±1.39}}
& 79.56 \\

GA
& 94.65\textsubscript{\textcolor{gray}{±1.04}}
& 92.75\textsubscript{\textcolor{gray}{±0.40}}
& 90.45\textsubscript{\textcolor{gray}{±0.72}}
& 53.76\textsubscript{\textcolor{gray}{±1.13}}
& 82.24\textsubscript{\textcolor{gray}{±1.00}}
& \redc{79.20\textsubscript{\textcolor{gray}{±2.83}}}
& 74.93\textsubscript{\textcolor{gray}{±3.12}}
& 81.14 \\

DE
& 93.29\textsubscript{\textcolor{gray}{±0.34}}
& \orangec{93.38\textsubscript{\textcolor{gray}{±0.19}}}
& 89.98\textsubscript{\textcolor{gray}{±0.24}}
& \orangec{55.25\textsubscript{\textcolor{gray}{±0.37}}}
& 82.18\textsubscript{\textcolor{gray}{±1.04}}
& 76.47\textsubscript{\textcolor{gray}{±0.38}}
& 73.08\textsubscript{\textcolor{gray}{±4.95}}
& 80.52 \\

GRACE 
& 93.61\textsubscript{\textcolor{gray}{±0.53}}
& 90.92\textsubscript{\textcolor{gray}{±1.15}}
& 89.60\textsubscript{\textcolor{gray}{±1.51}}
& \yellowc{53.96\textsubscript{\textcolor{gray}{±0.93}}}
& 82.34\textsubscript{\textcolor{gray}{±0.39}}
& 72.53\textsubscript{\textcolor{gray}{±8.62}}
& 73.92\textsubscript{\textcolor{gray}{±3.05}}
& 79.55 \\

PRL 
& \redc{96.32\textsubscript{\textcolor{gray}{±0.04}}}
& \yellowc{92.83\textsubscript{\textcolor{gray}{±0.24}}}
& \redc{91.27\textsubscript{\textcolor{gray}{±0.05}}}
& \redc{56.21\textsubscript{\textcolor{gray}{±0.15}}}
& 84.36\textsubscript{\textcolor{gray}{±0.08}}
& \yellowc{77.07\textsubscript{\textcolor{gray}{±2.36}}}
& \orangec{76.90\textsubscript{\textcolor{gray}{±0.95}}}
& \orangec{82.14} \\

\ours{}
& 95.35\textsubscript{\textcolor{gray}{±0.14}}
& 92.35\textsubscript{\textcolor{gray}{±0.05}}
& 90.57\textsubscript{\textcolor{gray}{±0.21}}
& 53.27\textsubscript{\textcolor{gray}{±0.66}}
& \yellowc{85.93\textsubscript{\textcolor{gray}{±0.62}}}
& \yellowc{77.07\textsubscript{\textcolor{gray}{±3.30}}}
& \yellowc{75.93\textsubscript{\textcolor{gray}{±0.40}}}
& \yellowc{81.50} \\

\ours{} (E)
& \orangec{95.88\textsubscript{\textcolor{gray}{±0.24}}}
& 92.55\textsubscript{\textcolor{gray}{±0.35}}
& \orangec{91.00\textsubscript{\textcolor{gray}{±0.65}}}
& 53.33\textsubscript{\textcolor{gray}{±0.35}}
& \redc{87.39\textsubscript{\textcolor{gray}{±0.35}}}
& \orangec{78.40\textsubscript{\textcolor{gray}{±1.22}}}
& \redc{80.98\textsubscript{\textcolor{gray}{±0.67}}}
& \redc{82.79} \\  \hline

\ours{} (I)
& 95.04\textsubscript{\textcolor{gray}{±0.18}}
& 91.53\textsubscript{\textcolor{gray}{±0.65}}
& 90.43\textsubscript{\textcolor{gray}{±0.21}}
& 49.79\textsubscript{\textcolor{gray}{±1.05}}
& 85.38\textsubscript{\textcolor{gray}{±0.20}}
& 74.33\textsubscript{\textcolor{gray}{±4.77}}
& 59.52\textsubscript{\textcolor{gray}{±2.29}}
& 78.00 \\

\ours{} (LOO)
& 95.70\textsubscript{\textcolor{gray}{±0.31}}
& 92.15\textsubscript{\textcolor{gray}{±0.15}}
& 90.42\textsubscript{\textcolor{gray}{±0.28}}
& 53.18\textsubscript{\textcolor{gray}{±1.40}}
& \orangec{86.43\textsubscript{\textcolor{gray}{±0.72}}}
& 75.87\textsubscript{\textcolor{gray}{±2.37}}
& 69.73\textsubscript{\textcolor{gray}{±3.68}}
& 80.50 \\

\hline
\end{tabular}%
}
\caption{Accuracy on classification tasks, averaged over three runs.
Colours mark the best (\redc{red}), second-best (\orangec{orange}) and third-best (\yellowc{yellow}) numbers in each column; minor differences ($\le{}0.05$) are treated as ties.
The right-most column shows the mean accuracy of each method across the seven datasets.}
\label{tab:cls}
\end{table*}

\subsection{Baselines}

\begin{itemize}
    \item \textbf{MI (Manual Instruction)}~\citep{zhang2022opt}: A set of prompts handcrafted and written by humans, aiming to improve task-specific performance.

    \item \textbf{NI (Natural Instruction)}~\citep{mishra2021cross}:
    Contains similarly to MI a set of human-written prompts for classification.

    \item \textbf{APE (Automatic Prompt Engineer)}~\citep{zhou2022large}: Automatically generates multiple instruction candidates with an LLM and selects the most effective prompt based on downstream performance, without further refinement during optimization. 
    This method only rephrases instructions and does not generate few-shot examples.

    \item \textbf{APO (Automatic Prompt Optimization)}~\citep{pryzant2023automatic}: Frames prompt tuning as a black-box optimization problem, refining prompts through an iterative feedback loop with beam search.
    Incorporate few-shot examples taken directly from the training dataset.

    \item \textbf{EvoPrompt}~\citep{guo12connecting}: Uses evolutionary strategies, selection, crossover, and mutation—to evolve a pool of discrete prompts and discover high-performing candidates. 
    Similar to APE, only rephrases instructions and does not generate few-shot examples.
    \begin{itemize}
        \item \textbf{DE (Differential Evolution)}: Explores the prompt space using differential evolution strategies.
        \item \textbf{GA (Genetic Algorithm)}: Applies genetic operators such as selection, crossover, and mutation to progressively improve prompt quality. 
    \end{itemize}
    
    \item \textbf{GRACE}~\citep{grace}: 
    Proposes a prompt optimization framework based on gated refinement and adaptive compression. 
    GRACE iteratively refines prompts while enforcing a no-regression constraint, only accepting modifications that do not degrade performance.
    To control prompt length and improve efficiency, it adaptively compresses prompts by removing or rewriting less useful components.

    \item \textbf{PRL (Prompts from Reinforcement Learning)}~\citep{batorski2025prl}: Employs a reinforcement learning framework to automatically generate and optimize prompts.
    PRL also constructs few-shot examples that are not in the training set.
    \item \textbf{\ours{}}: Our method as described in Section~\ref{sec:method}. 
    The first two variants \ours{} and \ours (E) are used throughout experiments, while the (I) and (LOO) variants are ablations. All variants otherwise have the same hyperparameters.
    \begin{itemize}
        \item \textbf{\ours{}}: With medium runtime budget with limited access to the training set.
        \item \textbf{\ours{} (E)}: With extended runtime budget and accessing the full dataset.
        \item \textbf{\ours{} (I)}: Use only the initially generated examples, without the replace/keep/drop cycle. Notably this variant does not access the training set.
        \item \textbf{\ours{} (LOO)}: Replace Shapley value selection~\eqref{eq:shapley} by simple leave-one-out.
    \end{itemize}
\end{itemize}


\subsection{Results}
\paragraph{Classification}
We evaluate \ours{} on seven standard text classification datasets spanning sentiment (SST-2, MR, CR, SST-5), question type (TREC), news topic (AG's News), and subjectivity (SUBJ). Results per--task are reported in Table~\ref{tab:cls} and summarized in Figure~\ref{fig:teaser}, including average runtime and data usage; per-dataset runtimes appear in Appendix~\ref{sec:runtime}. Overall, \ours{} is consistently among the top methods while being the fastest across benchmarks, and \ours{} (E) further improves performance, setting new state-of-the-art results on AG's News and SUBJ. Example prompts are provided in Appendix~\ref{sec:cls_prompt}.


\paragraph{Summarization}
We evaluate \ours{} on abstractive summarization using \textsc{SAMSum}~\citep{gliwa2019samsum}, a dataset of messenger-style dialogues with human-written summaries. We report ROUGE-1/2/L~\citep{lin2004rouge}, measuring unigram overlap (coverage), bigram overlap (local coherence), and longest common subsequence (fluency/structure). Table~\ref{tab:sum} shows that \ours{} is the fastest method and ranks second across all ROUGE metrics. An interesting observation is that, although PRL is capable of generating examples, it does not utilize any for the summarization task. Instead, PRL merely rephrases the manual prompt. The authors of PRL argue that summarization is not particularly suitable for example-based prompting. While we find that incorporating examples can indeed enhance performance, the improvements do not reach the level achieved by PRL. Finally, \ours{} (E) improves further, achieving the best ROUGE-2 and the best average over ROUGE-1/2/L. The \ours{} prompt is in Appendix~\ref{sec:sum_prompt}.

\begin{table}[ht]
\small
  \centering
  \begin{adjustbox}{max width=\columnwidth}
  \begin{tabular}{lcccc}
    \toprule
    \textbf{Method} & \textbf{ROUGE-1} & \textbf{ROUGE-2} & \textbf{ROUGE-L} & \textbf{Time [m]} \\
    \midrule
    MI                  & 32.76 & 10.39 & 28.97 & -- \\
    APE                 & 37.12\textsubscript{\textcolor{gray}{±2.02}} & 12.97\textsubscript{\textcolor{gray}{±0.74}} & 33.32\textsubscript{\textcolor{gray}{±1.68}} & \orangec{60.07}\textsubscript{\textcolor{gray}{±0.27}} \\
    GA                  & 39.69\textsubscript{\textcolor{gray}{±1.76}} & 14.47\textsubscript{\textcolor{gray}{±1.00}} & 35.84\textsubscript{\textcolor{gray}{±1.63}} & 89.31\textsubscript{\textcolor{gray}{±3.08}} \\
    DE                  & 33.91\textsubscript{\textcolor{gray}{±4.04}} & 12.53\textsubscript{\textcolor{gray}{±1.47}} & 31.05\textsubscript{\textcolor{gray}{±3.79}} & \yellowc{76.89}\textsubscript{\textcolor{gray}{±1.34}} \\
    
       GRACE                 & 40.61\textsubscript{\textcolor{gray}{±0.54}} & 14.65\textsubscript{\textcolor{gray}{±0.53}} & 35.86\textsubscript{\textcolor{gray}{±0.54}} &  1125.38\textsubscript{\textcolor{gray}{±21.40}} \\
    
    PRL                 & \redc{42.47\textsubscript{\textcolor{gray}{±0.83}}} & \orangec{16.17\textsubscript{\textcolor{gray}{±0.24}}} & \redc{37.73\textsubscript{\textcolor{gray}{±0.36}}} & \meanstd{2880.00}{0.00} \\
    \ours{}             & \yellowc{41.13}\textsubscript{\textcolor{gray}{±0.67}} & \yellowc{16.07}\textsubscript{\textcolor{gray}{±0.76}} & \yellowc{36.74}\textsubscript{\textcolor{gray}{±0.48}} & \redc{34.48}\textsubscript{\textcolor{gray}{±0.27}} \\
    \ours{} (E)         & \orangec{42.13}\textsubscript{\textcolor{gray}{±0.27}} & \redc{16.83}\textsubscript{\textcolor{gray}{±0.3}} & \orangec{37.37}\textsubscript{\textcolor{gray}{±0.25}} & 737.00\textsubscript{\textcolor{gray}{±108.31}} \\
    \bottomrule
  \end{tabular}
  \end{adjustbox}
      \caption{Text summarization results with ROUGE scores and runtime (minutes).}
  \label{tab:sum}
\end{table}

\paragraph{Domain knowledge task.}
To evaluate how \ours{} performs on tasks that require domain-specific knowledge, we additionally test it on MedQA~\citep{medqa}, a multiple-choice medical question answering benchmark with four options (A--D). Results are reported in Table~\ref{tab:medqa}. \ours{} achieves the second-best accuracy while being the fastest method overall, and \ours{} (E) attains only a small additional improvement despite a substantially higher runtime. This suggests that, for domain tasks, most benefits of refinement are realized within the first few iterations of the replace/drop/keep loop, after which gains diminish.

\begin{table}[ht]
\centering
\small
\renewcommand{\arraystretch}{1.2}
\begin{tabular}{lcc}
\hline
Method & Accuracy & Time[m] \\
\hline
APE   & 45.66\textsubscript{\textcolor{gray}{±0.97}} & \orangec{34.67\textsubscript{\textcolor{gray}{±9.75}}} \\
GA    & 51.95\textsubscript{\textcolor{gray}{±1.61}} & \yellowc{35.71\textsubscript{\textcolor{gray}{±2.73}}} \\
DE    & 51.76\textsubscript{\textcolor{gray}{±0.16}} & 88.63\textsubscript{\textcolor{gray}{±3.57}} \\
GRACE & 52.26\textsubscript{\textcolor{gray}{±0.16}} & 61.33\textsubscript{\textcolor{gray}{±9.74}} \\
PRL   & \redc{53.34\textsubscript{\textcolor{gray}{±0.11}}} & 2880.00\textsubscript{\textcolor{gray}{±0.00}} \\
\ours{}     & \yellowc{52.45\textsubscript{\textcolor{gray}{±0.51}}} & \redc{23.58\textsubscript{\textcolor{gray}{±0.61}}} \\
\ours{} (E) & \orangec{52.89\textsubscript{\textcolor{gray}{±0.05}}} & 617.55\textsubscript{\textcolor{gray}{±89.20}} \\ 
\hline
\end{tabular}
\caption{Results on MedQA dataset}
\label{tab:medqa}
\end{table}

\paragraph{Reasoning tasks}
We further evaluate \ours{} on mathematical reasoning benchmarks: GSM8K~\citep{cobbe2021gsm8k}, which requires explicit multi-step arithmetic with free-form integer answers, as well as DeepMath~\citep{deepmath} and Math500~\citep{math500}. Performance on reasoning tasks is known to be highly sensitive to the choice of in-context exemplars~\citep{wei2022chain}, making them a natural benchmark for example-centric prompt construction.
As shown in Table~\ref{tab:gsm8k}, methods that primarily rephrase or adjust the base instruction (APE, GA, DE) provide only modest improvements over the manual prompt, whereas methods that optimize few-shot examples (PRL, \ours{}) achieve substantially higher accuracy. \ours{} attains the strongest overall results and is also the fastest among the top-performing approaches, while \ours{} (E) yields a small additional gain at higher cost. Overall, these results show that \ours{} is both effective and compute-efficient for reasoning-intensive tasks.

\begin{table}
\centering
\begin{adjustbox}{max width=\columnwidth}
\begin{tabular}{lcccc}
\toprule
Method & GSM8K & DeepMath & MATH500 & Time[m] \\
\midrule
APE         & 83.43\textsubscript{\textcolor{gray}{±1.98}} & 15.47\textsubscript{\textcolor{gray}{±0.45}} & 31.53\textsubscript{\textcolor{gray}{±1.04}} & \orangec{180.81\textsubscript{\textcolor{gray}{±2.66}}} \\
GA          & 81.62\textsubscript{\textcolor{gray}{±1.38}} & 18.63\textsubscript{\textcolor{gray}{±2.37}} & 40.13\textsubscript{\textcolor{gray}{±1.39}} & \yellowc{191.96\textsubscript{\textcolor{gray}{±1.11}}} \\
DE          & 79.52\textsubscript{\textcolor{gray}{±0.45}} & 16.10\textsubscript{\textcolor{gray}{±0.00}} & 34.20\textsubscript{\textcolor{gray}{±1.39}} & 252.57\textsubscript{\textcolor{gray}{±3.59}} \\
GRACE       & 82.37\textsubscript{\textcolor{gray}{±1.82}} & 15.05\textsubscript{\textcolor{gray}{±0.16}} & 33.20\textsubscript{\textcolor{gray}{±1.60}} & 1436.41\textsubscript{\textcolor{gray}{±74.30}} \\
PRL         & \yellowc{86.15\textsubscript{\textcolor{gray}{±0.55}}} & \yellowc{21.58\textsubscript{\textcolor{gray}{±0.22}}} & \yellowc{44.40\textsubscript{\textcolor{gray}{±1.40}}} & 2880.00\textsubscript{\textcolor{gray}{±0.00}} \\
\ours{}     & \orangec{91.65\textsubscript{\textcolor{gray}{±0.31}}} & \orangec{24.85\textsubscript{\textcolor{gray}{±1.13}}} & \orangec{48.53\textsubscript{\textcolor{gray}{±0.31}}} & \redc{80.26\textsubscript{\textcolor{gray}{±2.95}}} \\
\ours{} (E) & \redc{92.12\textsubscript{\textcolor{gray}{±0.12}}} & \redc{25.33\textsubscript{\textcolor{gray}{±0.75}}} & \redc{48.70\textsubscript{\textcolor{gray}{±0.50}}} & 1598.34\textsubscript{\textcolor{gray}{±234.54}} \\
\bottomrule
\end{tabular}
\end{adjustbox}
\captionof{table}{Results on GSM8K, DeepMath, and MATH500.}
\label{tab:gsm8k}
\end{table}

\subsection{Ablations}

\paragraph{Ablation Study: Cross-Model Robustness}
We assess how well prompts learned by \ours{} transfer across models in two settings. 
\textbf{(i) Cross-model inference.} We train prompts with Qwen2.5-7B-Instruct and then evaluate the same prompts using Mistral-7B-Instruct-v0.2~\citep{jiang2023mistral7b}.
As shown in Table~\ref{tab:subj_robustness} (left), \ours{} attains the strongest transfer on \textsc{SUBJ}, edging out PRL and APO; APE and GA are roughly on par with the manual instruction baseline.
Notably, \ours{} (E) exhibits a sizable accuracy drop, which we attribute to overfitting to the source evaluator due to its substantially larger improvement iteration budget.
These results suggest that, when portability matters, \ours{} offers the best cross-model robustness. 
\textbf{(ii)~Component swaps.} We next vary which model plays each role, swapping Qwen and Mistral between the (Example Proposer \& Improver) and the Prompt Evaluator. Table~\ref{tab:subj_robustness} (right) shows that accuracy remains comparable across configurations, indicating that \ours{} is not overly sensitive to a particular model pairing.

    
    

\begin{table}[ht]
\small
\centering
\vspace{-0.6em}

\begin{minipage}[t]{0.45\columnwidth}
\centering
\begin{adjustbox}{max width=\linewidth}
\begin{tabular}{l c}
  \toprule
  \textbf{Method} & \textbf{Acc.} \\
  \midrule
  MI              & 60.30 \\
  APE             & 60.77\textsubscript{\textcolor{gray}{±1.08}} \\
  APO             & \yellowc{69.53\textsubscript{\textcolor{gray}{±1.33}}} \\
  GA              & 60.68\textsubscript{\textcolor{gray}{±1.60}} \\
  DE              & 64.10\textsubscript{\textcolor{gray}{±2.20}} \\
  GRACE           & 69.10\textsubscript{\textcolor{gray}{±1.88}} \\
  PRL             & \orangec{70.73\textsubscript{\textcolor{gray}{±3.81}}} \\
  \ours{}         & \redc{72.87\textsubscript{\textcolor{gray}{±4.16}}} \\
  \ours{} (E)     & 68.75\textsubscript{\textcolor{gray}{±3.01}} \\
  \bottomrule
\end{tabular}
\end{adjustbox}
\end{minipage}
\hfill
\begin{minipage}[t]{0.5\columnwidth}
\centering
\begin{adjustbox}{max width=\linewidth}
\begin{tabular}{l c c}
  \toprule
  \textbf{P \& I} & \textbf{Eval} & \textbf{Acc.} \\
  \midrule
  Qwen    & Mistral & \meanstd{75.93}{3.14} \\
  Mistral & Qwen    & \meanstd{72.52}{1.71} \\
  Mistral & Mistral & \meanstd{74.93}{2.54} \\
  \bottomrule
\end{tabular}
\end{adjustbox}
\end{minipage}

\caption{Cross-model robustness on \textsc{SUBJ}. \textbf{Left:} prompts trained with Qwen2.5-7B-Instruct, evaluated with Mistral-7B-Instruct-v0.2. \textbf{Right:} accuracy under component swaps between Qwen and Mistral (Example Proposer \& Improver vs.\ Prompt Evaluator).}
\label{tab:subj_robustness}
\vspace{-0.6em}
\end{table}

\paragraph{Ablation Study: Influence of the Replace/Drop/Keep Optimization}

In this experiment, we evaluate \ours{} without the optimization loop, i.e., the model is tested directly on the proposed initial examples. 
We include this variant as a baseline, denoted \textsc{\ours{} (I)}, in Table~\ref{tab:cls}. 
As shown, for many benchmarks the initial examples already yield strong performance, in some cases even surpassing algorithms that employ optimization loops. 
Nevertheless, we consistently observe that incorporating our optimization loop further improves the results. 
For certain tasks, such as binary sentiment classification (e.g., SST-2 or MR), the improvement is marginal. 
We attribute this to the fact that the initial examples are already highly effective, as the underlying LLM has a strong capability to distinguish between positive and negative samples. 

\noindent 
Interestingly, in the subjectivity dataset \ours{} without the optimization loop performs poorly, achieving results comparable to those of the Manual Instruction baseline. 
However, after applying the optimization loop, performance improves significantly by 16.41\% and by an additional 5.05\% when using \ours (E), showing that on more difficult tasks where in-context example selection is non-trivial our optimization loop can help a lot.

\paragraph{Ablation Study: Influence of \#Replace/Drop/Keep Iterations} 
\label{sec:craft}

\begin{figure}
  \centering
  \includegraphics[width=\linewidth]{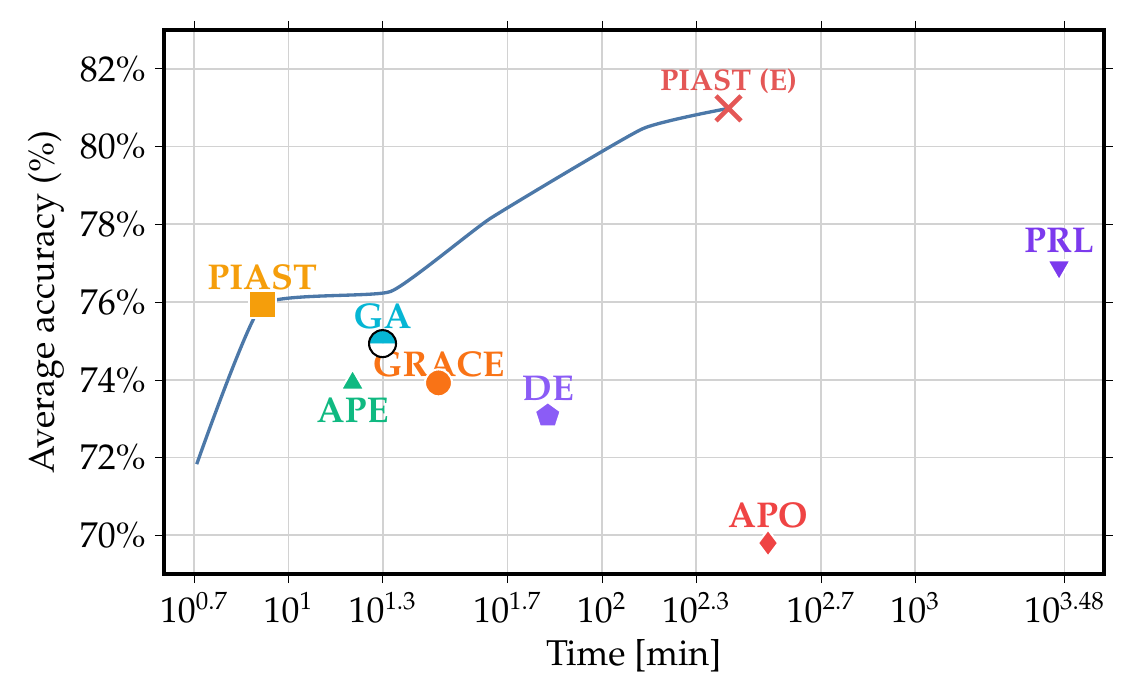}
  \caption{Scaling of \ours{} on SUBJ compared to other baselines while increasing the number of improvement iterations.}
  \label{fig:abl_subj}
\end{figure}

We investigate how \ours{} scales with an increasing number of crafting iterations, 
and compare it against baselines from the literature. 
To this end, we run \ours{} with 10, 15, 30, 50, 100, and 150 crafting iterations, 
and report the results in Figure~\ref{fig:abl_subj}. 
\noindent 
We observe a clear trend: increasing the number of crafting iterations consistently 
improves accuracy, albeit at the cost of higher runtime. 
This highlights an appealing property of \ours{}: its performance can be effectively scaled by allocating more computation time by increasing the number of crafting iterations. 
Moreover, the plot clearly shows that \ours{} has anytime performance superior to the baselines.

\paragraph{Ablation: Leave–One–Out vs.\ Shapley for worst–example selection}
To test whether a simpler procedure can replace our Shapley–value selection, we evaluate a
\emph{leave–one–out} (LOO) heuristic for identifying the worst (most harmful) in–context example.
Given $n$ examples $E=\{e_i\}_{i=1}^n$, LOO removes each example
once and measures performance on the validation split:
\[
i^\star \;=\; \argmax_{i\in\{1,\dots,N\}} \; v\!\left(E\setminus\{e_i\}\right).
\]
In words, LLO chooses the example whose removal yields the highest accuracy drop.
removing a strongly useful example would decrease it.

\noindent 
We run this ablation on all classification tasks using the same hyperparameters as \ours{}
and report results in Table~\ref{tab:cls}.
Across several benchmarks LOO attains results comparable to our full method, but on \textsc{Subj} there is a clear gap between \ours{} and \ours{} (LOO).
We hypothesize that, for many classification tasks, the initial pool already contains mostly good examples, so LOO can make small, beneficial adjustments.
In contrast, \textsc{Subj} is more sensitive to initialization (see \ours{} (I)), and the Shapley–based selection is notably more robust in such settings, where performance depends heavily on which examples are initially presented.


\section{Conclusions}
\label{sec:conclusion}
\noindent 
We have shown \ours{}, a fast and effective prompting method that synthesizes ICL examples and outperforms several sota methods on a wider range of benchmarks.
\noindent 
Interestingly, we argue that \ours{} relies less on the training set than some other approaches. The only information it gets from the training set is its accuracy with the given prompt. Other approaches often use more information, e.g.\ using examples from the test set in the prompts. Additionally, we show that we do not need the full training set to achieve high performance, a subset typically suffices. We conjecture \ours{} is a step towards ever less reliance on training sets for prompting.
\noindent 
\ours{} relies on the LLM having some task-specific competence to propose useful examples. In cases where this is not the case, a more involved iterative example synthesis might be needed.
\noindent
We also believe that, while example quality is often the dominant lever, combining \ours{} with instruction rewriting could yield complementary gains. 
\noindent 
Another promising direction is to make the proposer benefit from other external signals (e.g., retrieval from a small seed set, weak heuristics, or verification-style self-checks) or by using a stronger or more specialized model for proposal while keeping the evaluator lightweight.


\section{Limitations}
\label{sec:limitations}
Performance and applicability of \ours{} are limited by:
\begin{description}[leftmargin=!, labelwidth=10pt]
    \item[Task-specific competence] The LLM having task-specific competence: If the LLM cannot propose meaningful examples at all, the final prompt is unlikely to have helpful ICL examples.
    \item[Training set] We rely on a training set to evaluate our ICL examples with. If no training set is given, PIAST can still propose initial ICL examples, but will not be able to improve them.
    \item[Scoring function] Second, we need a scoring function on the training set as well. When it is not aligned with the ultimate task performance, prompts generated with \ours{} might not help.
    \item[Prompt construction overhead] \ours{} incurs an initial overhead while constructing the prompt. While significantly smaller than other competing methods, it still takes a few minutes. This cost is incurred only once, however, and the generated ICL examples can be reused for any following task prompt.
    \item[Larger \& proprietary LLMs] While we have tested on state of the art medium sized open-source models, performance on larger and proprietary models might differ.
\end{description}

\section*{Ethics Statement}
This work was conducted in accordance with the ACL Code of Ethics and the ACM Code of Ethics and Professional Conduct, and we have adhered to the rules throughout this research. We recognize that large pre-trained language models, whether used with automated prompting or not, have the potential to be misused to generate misleading, toxic, or otherwise harmful content. Our intention is that the proposed prompting method contributes to improving the steerability of such models and supports more responsible use. We also acknowledge that appropriate safeguards and further evaluation are necessary to mitigate risks in real world applications.

\bibliography{custom}

\newpage 
\appendix

\section{Additional Experimental Result - Simplification}
\label{sec:sim}
We evaluate \ours{} on the sentence simplification task using the \textsc{ASSET} dataset \citep{alva2020asset}. 
ASSET is a crowdsourced corpus specifically designed to evaluate simplification models across a range of rewriting operations, including lexical paraphrasing, sentence splitting, deletion, and reordering. 
Each original sentence is paired with multiple human-written simplifications.

For measuring simplification quality, we adopt the SARI metric \citep{xu2016optimizing}, which compares the system output to both the original sentence and the reference simplifications.
Results are presented in Table~\ref{tab:sim}. As shown, \ours{} 
achieves the highest SARI score for text simplification while 
requiring the least runtime. Furthermore, \ours{} exhibits the 
lowest standard deviations across runs, highlighting its stability 
and robustness. With additional computational time and data, 
\ours{} (E) attains an even higher score on this benchmark. The example prompt is given in Appendix \ref{sec:sim_prompt}.

\begin{table}[ht]
\centering
\small
\begin{adjustbox}{max width=\columnwidth}
\begin{tabular}{lcc}
\toprule
\textbf{Method} & \textbf{SARI} & \textbf{Time[m]}  \\
\midrule
MI              & 43.77 & -- \\
APE             & 45.33\textsubscript{\textcolor{gray}{±0.83}} & \orangec{35.69\textsubscript{\textcolor{gray}{±0.20}}} \\
GA & 46.25\textsubscript{\textcolor{gray}{±0.47}} & \yellowc{39.60\textsubscript{\textcolor{gray}{±0.63}}} \\
DE  & 45.79\textsubscript{\textcolor{gray}{±0.35}} & 52.77\textsubscript{\textcolor{gray}{±1.12}} \\
GRACE  & 50.21\textsubscript{\textcolor{gray}{±0.18}}& 611.61\textsubscript{\textcolor{gray}{±19.57}} \\
PRL             & \yellowc{52.26}\textsubscript{\textcolor{gray}{±3.51}} & 2880.00\textsubscript{\textcolor{gray}{±0.00}} \\
\ours{}         & \orangec{54.52\textsubscript{\textcolor{gray}{±0.07}}} & \redc{18.14\textsubscript{\textcolor{gray}{±0.42}}} \\

\ours{} (E)        & \redc{55.06\textsubscript{\textcolor{gray}{±0.26}}} & 389.78\textsubscript{\textcolor{gray}{±113.17}} \\
\bottomrule
\end{tabular}
\end{adjustbox}
\caption{Simplification task results.}
\label{tab:sim}
\end{table}

\section{Hyperparameters}
\label{sec:hyperparameters}

In this section, we present the exact hyperparameters used across all tasks. 
A single set of hyperparameters was applied consistently across all tasks, and their values 
are summarized in Table~\ref{tab:hyperparameters}.

\begin{table}[h]
\centering
\begin{tabular}{lc}
\toprule
\textbf{Hyperparameter} & \textbf{Value} \\
\midrule
$k$ & 16 \\
$s$ & 70 \\
$I$ & 15 \\
$m$ & 10 \\
$r$ & 5  \\
$P$ & 3  \\
\bottomrule
\end{tabular}
\caption{Hyperparameters used across all tasks. The notation is consistent with the pseudocode provided in Appendix~\ref{sec:pseudocode}.}
\label{tab:hyperparameters}
\end{table}

The hyperparameters for \ours{} (E) are identical to those of \ours{}, except that we increase the number of iterations $I$ to 150. 
This adjustment provides the most effective way to scale the performance of \ours{}, as demonstrated in our ablation studies (see Section~\ref{sec:craft}).

\section{Hyperparameter Study}
We study the impact of the number of Shapley permutations, number of refine and initial example candidates for performance of \ours{}.

\begin{table}[ht]
\centering
\begin{tabular}{@{}rcc@{}}
\toprule
\textbf{m} & \textbf{Acc.} & \textbf{Time (min)} \\
\midrule
5   & \meanstd{70.57}{2.33} & \meanstd{ 7.11}{0.15} \\
10   & \meanstd{75.93}{0.07} & \meanstd{8.25}{0.27} \\
30   & \meanstd{76.15}{0.48} &  \meanstd{11.54}{0.05}\\
50  & \meanstd{76.33}{1.53} &  \meanstd{12.99}{0.32} \\
\bottomrule
\end{tabular}
\caption{Effect of the number of proposed candidates on the subjectivity task.}
\label{tab:ablation_craft_iterations}
\end{table}

\paragraph{Hyperparameter Study: Influence of Number of Refine Candidates} 
In this experiment, we analyze how the number of candidate examples $m$ proposed by the example improver influences the final performance. 
The results are summarized in Table~\ref{tab:ablation_craft_iterations}. 

We observe that using too few candidates ($m=5$) leads to a significant drop in accuracy, which can be mitigated by increasing the number of candidates to $m=10$. 
For larger values $m=30$ and $m=50$, accuracy does not improve substantially, suggesting that an initial pool of $m=10$ candidates combined with the replace/drop/keep iteration is sufficient to achieve strong performance. 
Therefore, in all subsequent experiments we adopt $m=10$ as the default setting.

\paragraph{Hyperparameter Study: Influence of Number of Initial Examples}

In this experiment, we investigate how performance on the subjectivity task varies with the number of initial examples. 
The results are reported in Table \ref{tab:ablation_initial}. 
The number of initial examples does not exhibit a clear monotonic trend (e.g., ``the more, the better''). 
Using too few initial examples may lead to underfitting, as they fail to provide sufficient insight into the task. 
Conversely, using too many examples can introduce excessive noise, making it more difficult for the evaluator to identify the weakest examples within a large pool. 
Based on this analysis, we select 16 examples as the most robust choice for our experiments.
\begin{table}[ht]
\centering
\begin{tabular}{@{}rcc@{}}
\toprule
\textbf{k} & \textbf{Acc.} & \textbf{Time [m]} \\
\midrule
4  & \meanstd{71.77}{0.42} & \meanstd{2.88}{0.12} \\
8  & \meanstd{74.20}{1.66} & \meanstd{4.63}{0.14} \\
16 & \meanstd{75.93}{0.40} & \meanstd{8.25}{0.27} \\
32 & \meanstd{74.77}{0.48} & \meanstd{16.26}{0.26} \\
\bottomrule
\end{tabular}
\caption{Effect of the number of initial examples on the subjectivity task.}
\label{tab:ablation_initial}
\end{table}

\paragraph{Hyperparameter Study: Influence of Number of Shapley Permutations}
We study the sensitivity of \ours{} to the number of Monte--Carlo permutations $K$ used in the Shapley-value estimator~\eqref{eq:mc-shapley}.
As shown in Table~\ref{tab:ablation_shap}, increasing $K$ beyond $3$ yields only marginal accuracy gains (at most $0.39$ points from $K = 3$ to $K=50$) while substantially increasing runtime. Using $K=1$ underperforms $K=3$ by $2.25$ points, indicating that a small amount of averaging is beneficial. We therefore adopt $K=3$ as our default, which offers a strong performance/speed trade-off. 

\begin{table}
\small
\centering
\begin{adjustbox}{max width=\columnwidth}
\begin{tabular}{@{}rcc@{}}
\toprule
\textbf{P} & \textbf{Acc.} & \textbf{Time [m]} \\
\midrule
1  & \meanstd{73.68}{1.45} & \meanstd{4.61}{0.04} \\
3  & \meanstd{75.93}{0.40} & \meanstd{8.25}{0.27} \\
10 & \meanstd{76.02}{1.08} & \meanstd{20.88}{0.17} \\
50 & \meanstd{76.32}{1.41} & \meanstd{83.90}{1.82} \\
\bottomrule
\end{tabular}
\end{adjustbox}
\caption{Effect of the number of Shapley permutations on subjectivity.}
\label{tab:ablation_shap}
\end{table}

\section{Runtime Analysis for Classification Benchmarks}
\label{sec:runtime}

In this section, we report the average runtime (in minutes) across different methods for each classification benchmark. 
Table~\ref{tab:runtime_mean_sd} summarizes the results, including mean and standard deviation values.

\begin{table*}[ht]
\centering
\resizebox{\linewidth}{!}{%
\begin{tabular}{lcccccccc}
\toprule
Dataset & APE & APO & DE & GA & GRACE & PRL & \ours{} & \ours{} (E) \\
\midrule
sst2   & \meanstd{62.85}{1.96} & \meanstd{376.82}{0.00} & \meanstd{62.79}{2.54} & \meanstd{20.21}{2.84} & \meanstd{28.76}{1.48} & \meanstd{2880.00}{0.00} & \meanstd{7.49}{0.14} & \meanstd{221.48}{8.41}\\
cr     & \meanstd{16.09}{5.29} & \meanstd{302.22}{47.40} & \meanstd{65.06}{0.96} & \meanstd{18.75}{4.37} & \meanstd{28.87}{1.20} & \meanstd{2880.00}{0.00} & \meanstd{7.49}{0.20} & \meanstd{181.06}{8.67}\\
mr     & \meanstd{4.60}{0.45}  & \meanstd{342.03}{15.04} & \meanstd{62.71}{1.36} & \meanstd{28.84}{9.18} & \meanstd{27.88}{1.70} & \meanstd{2880.00}{0.00} & \meanstd{7.64}{0.11} & \meanstd{210.40}{20.84}\\
sst5   & \meanstd{5.99}{1.06}  & \meanstd{430.08}{92.80} & \meanstd{62.09}{1.88} & \meanstd{20.68}{1.82} & \meanstd{44.94}{14.45} & \meanstd{2880.00}{0.00} & \meanstd{7.63}{0.11} & \meanstd{155.55}{0.35}\\
agnews & \meanstd{7.33}{2.43}  & \meanstd{241.19}{28.07} & \meanstd{65.16}{3.59} & \meanstd{18.08}{2.54} & \meanstd{32.53}{4.09} & \meanstd{2880.00}{0.00} & \meanstd{8.38}{0.16} & \meanstd{210.97}{38.85}\\
trec   & \meanstd{3.94}{0.74}  & \meanstd{256.34}{17.55} & \meanstd{66.01}{3.25} & \meanstd{22.24}{6.46} & \meanstd{37.52}{2.75} & \meanstd{2880.00}{0.00} & \meanstd{6.63}{0.19} & \meanstd{ 146.19}{28.15}\\
subj   & \meanstd{16.03}{2.63} & \meanstd{339.44}{42.38} & \meanstd{67.28}{0.37} & \meanstd{19.98}{2.33} & \meanstd{30.14}{4.26} & \meanstd{2880.00}{0.00} & \meanstd{8.25}{0.27} & \meanstd{253.83}{8.07}\\
\bottomrule
\end{tabular}%
}
\caption{Average runtime in minutes (mean $\pm$ SD) for each classification benchmark and method. The last row reports the overall average across all APO runs.}
\label{tab:runtime_mean_sd}
\end{table*}

\section{Prompt for Example Generator}
\label{sec:prompt_generator}
In this section we give the prompt for the prompt evaluator for binary sentimental analysis task. For other tasks, the prompt is analogous.  

\promptbox{Example Proposer prompt for CR dataset}{You are a data generator that writes high-quality in-context learning examples for \emph{binary sentiment} on short movie-review style snippets. Create exactly \{NUM\_EXAMPLES\} training examples in THIS STRICT format only:\\[0.5em]
Example1:\\
Sentence: "``\textless text\textgreater{}''"\\
Label: \{LABEL\}\\[0.4em]
Example2:\\
Sentence: "``\textless text\textgreater{}''"\\
Label: \{LABEL\}\\[0.4em]
\ldots\\
Example\{NUM\_EXAMPLES\}:\\
Sentence: "``\textless text\textgreater{}''"\\
Label: \{LABEL\}\\[0.6em]
Diversity plan (MUST FOLLOW):\\
\{DIVERSITY\_PLAN\}\\[0.6em]
Rules:\\
- Each example's "Sentence" must contain exactly the number of sentences specified above (1--3).\\
- Keep sentences concise: typically 3--14 words each. Across the set, include at least one very short ($\leq$ 5 words) and one longer (10--14 words).\\
- Use only ASCII characters. Do NOT include double quotes inside the text.\\
- Use exactly ONE `Sentence:` line per example; if multiple sentences are needed, put them inside the same quotes separated by a space.\\
- Make the writing naturally match the requested label in the everyday sense of the word.\\
- Do NOT mention the label or talk about labels in the text (no meta commentary).\\
- No Markdown/code fences.\\
- Output ONLY the examples in the exact format above; no extra text.%
}{teal}

\section{Prompt for Example Improver}
\label{sec:prompt_improver}

\promptbox{Example Improver prompt for CR dataset}{%
You are improving in-context examples for sentiment classification. Generate replacements that diversify length (1--3 sentences) and topic, avoid paraphrasing, and help the task.\\[0.5em]
You are given the CURRENT examples (do not repeat or paraphrase them):\\
\{CURRENT\_EXAMPLES\}\\[0.6em]
Now create exactly \{NUM\_CANDIDATES\} NEW examples in THIS STRICT format:\\[0.2em]
Example1:\\
Sentence: "``\textless text\textgreater{}''"\\
Label: positive\textbar{}negative\\[0.4em]
Example2:\\
Sentence: "``\textless text\textgreater{}''"\\
Label: positive\textbar{}negative\\[0.4em]
\ldots\\
Example\{NUM\_CANDIDATES\}:\\
Sentence: "``\textless text\textgreater{}''"\\
Label: positive\textbar{}negative\\[0.6em]
Diversity plan (MUST FOLLOW):\\
\{DIVERSITY\_PLAN\}\\[0.6em]
Rules:\\
- Use exactly ONE `Sentence:` line per example. If multiple sentences are needed, put them INSIDE the same quotes separated by a space.\\
- Each example must have exactly the number of sentences specified in the plan above (1--3).\\
- Keep sentences concise: typically 3--14 words each. Across the set, include very short ($\leq$ 5 words) and longer (10--14 words).\\
- ASCII only. Do NOT include double quotes inside the text.\\
- Make topics clearly different from the given examples and from each other; avoid near-duplicates or paraphrases.\\
- Prefer balancing labels; if unsure, choose the minority label: \{MINORITY\_LABEL\}.\\
- Do NOT wrap output in Markdown/code fences.\\
- Output ONLY the examples in the exact format above; no extra text.%
}{violet}

\section{Classification Prompts} 
\label{sec:cls_prompt}

In this section, we describe the most effective prompts for \ours{} on classification tasks. 
The base prompts are taken from \citep{guo12connecting}, and the examples are produced by our method.

\promptbox{SST2}{Please perform Sentiment Classification task. Given the sentence, assign a sentiment label from [’negative’,  ’positive’]. Return label only without any other text.

  Example1:
  Sentence: "The film maintains a steady pace. No dull moments. Engaging from beginning to end."
  Label: positive

  Example2:
  Sentence: "The set design is detailed. Costumes match the era perfectly. Attention to historical accuracy is evident."
  Label: positive

  Example3:
  Sentence: "The lead actor delivers a compelling performance."
  Label: positive

  Example4:
  Sentence: "The film maintains a brisk pace from start to finish. No lulls or wasted moments, just continuous action and intrigue."
  Label: positive

  Example5:
  Sentence: "The soundtrack is loud and distracting, overshadowing the dialogue."
  Label: negative

  Example6:
  Sentence: "The editing is seamless, keeping the pace tight. Transitions between scenes are smooth and impactful."
  Label: positive

  Example7:
  Sentence: "The lead actor's performance is powerful and moving."
  Label: positive

  Example8:
  Sentence: "The cinematography is breathtaking, capturing the essence of the story. The use of light and color enhances every scene."
  Label: positive

  Example9:
  Sentence: "The lead actress's performance is powerful."
  Label: positive

  Example10:
  Sentence: "The direction was confusing and lacked focus."
  Label: negative

  Example11:
  Sentence: "The lead actress shines in every scene."
  Label: positive

  Example12:
  Sentence: "The dialogue felt forced and unnatural."
  Label: negative

  Example13:
  Sentence: "The visual effects were poorly done. The CGI looked cheap and unrealistic. It ruined the immersion."
  Label: negative

  Example14:
  Sentence: "The director masterfully guides the narrative."
  Label: positive

  Example15:
  Sentence: "The lead actress's portrayal is emotionally resonant."
  Label: positive}{teal}
\promptbox{CR}{
Please perform Sentiment Classification task. Given the sentence, assign a sentiment label from [’negative’,  ’positive’]. Return label only without any other text.

Example1:
  Sentence: "The cinematography captured the essence of the setting, with stunning visuals that added depth to the story."
  Label: positive

  Example2:
  Sentence: "The editing was precise, enhancing the flow of the story. However, some transitions felt abrupt and jarring."
  Label: negative

  Example3:
  Sentence: "The actors delivered powerful performances, bringing depth to their roles."
  Label: positive

  Example4:
  Sentence: "The screenplay was weak, with dialogue that felt forced and unnatural. Characters lacked development, making their actions confusing. The dialogue felt stiff, with lines that didn't flow naturally. This made the scenes less engaging."
  Label: negative

  Example5:
  Sentence: "The first act was slow and     \begin{CJK}{UTF8}{gbsn}
    拖沓，但中间部分节奏加快，保持了紧张感。
    \end{CJK}"
  Label: positive

  Example6:
  Sentence: "The editing was seamless, enhancing the flow of the story. Cuts were precise, keeping the pacing tight."
  Label: positive

  Example7:
  Sentence: "The film started slowly but picked up in the middle."
  Label: negative

  Example8:
  Sentence: "The director's vision was clear but the execution was lacking. Scenes felt disjointed, and the overall story was confusing."
  Label: negative

  Example9:
  Sentence: "The editing was choppy and disjointed."
  Label: negative

  Example10:
  Sentence: "The director's vision was clear, but the actors seemed uncomfortable on camera."
  Label: negative

  Example11:
  Sentence: "The screenplay felt rushed, with dialogue that seemed out of place. Characters had little to no development, making their motivations unclear. The plot relied too heavily on clichés, lacking originality."
  Label: negative

  Example12:
  Sentence: "The director's vision was clear and inspiring. However, the final cut felt rushed and incomplete."
  Label: negative

  Example13:
  Sentence: "The soundtrack was inappropriate, detracting from the mood of the scenes."
  Label: negative

  Example14:
  Sentence: "The director skillfully guided the ensemble cast."
  Label: positive

  Example15:
  Sentence: "The screenplay felt rushed, with dialogue that seemed out of place. Characters had little to no development, making their motivations unclear. The plot relied too heavily on clichés, lacking originality."
  Label: negative}{teal}
\promptbox{MR}{Please perform Sentiment Classification task. Given the sentence, assign a sentiment label from [’negative’,  ’positive’]. Return label only without any other text.

Example1:
  Sentence: "The editing was choppy, disrupting the flow of the narrative. Scenes felt disjointed, and the timing was off."
  Label: negative

  Example2:
  Sentence: "The lead actress delivered a nuanced and emotionally rich performance."
  Label: positive

  Example3:
  Sentence: "The movie started slowly but picked up momentum. The second act was particularly well-paced, maintaining tension."
  Label: positive

  Example4:
  Sentence: "The lead actor's performance was powerful and moving."
  Label: positive

  Example5:
  Sentence: "The lead actress captivated the audience with her portrayal."
  Label: positive

  Example6:
  Sentence: "The soundtrack was overbearing and distracting."
  Label: negative

  Example7:
  Sentence: "The soundtrack added a melancholic tone that complemented the film's somber mood."
  Label: positive

  Example8:
  Sentence: "The actor's portrayal was compelling and emotionally resonant."
  Label: positive

  Example9:
  Sentence: "The editing was choppy and disjointed."
  Label: negative

  Example10:
  Sentence: "The lead actor brought depth to the role."
  Label: positive

  Example11:
  Sentence: "The lead actor's portrayal was gripping and heartfelt."
  Label: positive

  Example12:
  Sentence: "The soundtrack added a perfect touch, enhancing the film's dramatic moments."
  Label: positive

  Example13:
  Sentence: "The visual effects were poorly done and noticeable."
  Label: negative

  Example14:
  Sentence: "The director's vision was unclear, leading to a disjointed narrative. The characters felt underdeveloped."
  Label: negative

  Example15:
  Sentence: "The acting was wooden and unconvincing."
  Label: negative}{teal}
\promptbox{SST5}{Please perform Sentiment Classification task. Given the sentence, assign a sentiment label from [’terrible’,  ’bad’, ’okay’, ’good’, ’great’]. Return label only without any other text.

  Example1:
  Sentence: "The screenplay was predictable. The dialogue lacked depth, feeling forced. Characters felt flat and uninteresting."
  Label: bad

  Example2:
  Sentence: "The story lacked coherence and felt rushed. The plot had too many loose ends and felt unsatisfying."
  Label: bad

  Example3:
  Sentence: "The screenplay was cliché and predictable. The dialogue lacked depth and felt forced."
  Label: bad

  Example4:
  Sentence: "The story was predictable with a weak ending. It lacked the twists needed for a compelling narrative."
  Label: okay

  Example5:
  Sentence: "The screenplay was clever and witty. The dialogue was sharp and well-paced. It captured the essence of the characters perfectly."
  Label: great

  Example6:
  Sentence: "Direction felt disjointed and confusing."
  Label: terrible

  Example7:
  Sentence: "Story lacked coherence and felt rushed."
  Label: terrible

  Example8:
  Sentence: "Dialogue was forced and awkward, breaking the mood."
  Label: terrible

  Example9:
  Sentence: "Direction kept the pace just right; not too slow or fast."
  Label: good

  Example10:
  Sentence: "The dialogue was cliché and predictable. The script failed to deliver any surprises."
  Label: okay

  Example11:
  Sentence: "The director skillfully balanced the dramatic and comedic elements. The pacing was just right, keeping the audience engaged. The visual storytelling was top-notch."
  Label: great

  Example12:
  Sentence: "Dialogue was sharp and added depth to the characters."
  Label: good

  Example13:
  Sentence: "The acting was solid but not memorable. The supporting cast added some depth to the film."
  Label: okay

  Example14:
  Sentence: "Acting was wooden and unconvincing."
  Label: terrible

  Example15:
  Sentence: "Great performances by all; especially the lead actor."
  Label: good

  Example16:
  Sentence: "The acting was wooden and unconvincing."
  Label: bad

  Example17:
  Sentence: "The direction felt disjointed and confusing."
  Label: bad

  Example18:
  Sentence: "The direction was uninspired. The pacing felt slow and the camera work was basic."
  Label: okay

}{teal}
\promptbox{AG's News}{
Please perform News Classification task. Given the news item, assign a label from [’World’, ’Sports’,  ’Business’, ’Tech’]. Return label only without any other text

Example1:
  Sentence: "Upcoming elections will focus on healthcare reform and immigration policies; debates intensify."
  Label: World

  Example2:
  Sentence: "NASA launches Mars rover to study planet's geology."
  Label: Tech

  Example3:
  Sentence: "Supreme Court rules on new labor laws; impact on businesses discussed."
  Label: Business

  Example4:
  Sentence: "Elections this year will focus on healthcare and education reform."
  Label: World

  Example5:
  Sentence: "Telemedicine platforms see surge in usage during pandemic."
  Label: Tech

  Example6:
  Sentence: "US and China engage in summit talks to discuss trade and climate issues; tensions remain high."
  Label: Business

  Example7:
  Sentence: "Vaccination drive reaches remote villages successfully."
  Label: World

  Example8:
  Sentence: "Diplomatic talks on climate change continue with mixed progress."
  Label: World

  Example9:
  Sentence: "Bombing kills dozens in city center."
  Label: World

  Example10:
  Sentence: "Upcoming elections will focus on healthcare and economic reforms; debates intensify as candidates present their plans."
  Label: World

  Example11:
  Sentence: "Security forces respond to a terrorist attack in the city center; multiple casualties reported. Emergency services work to contain the situation."
  Label: World

  Example12:
  Sentence: "Diplomatic talks on trade agreements between Asia-Pacific nations continue."
  Label: World

  Example13:
  Sentence: "Upcoming elections will focus on healthcare and economic reforms; debates intensify as candidates present their plans. Voters express concerns about rising costs."
  Label: World

  Example14:
  Sentence: "Diplomatic talks between nations on climate change progress despite initial disagreements."
  Label: World}{teal}
\promptbox{TREC}{Please perform Question Classification task. Given the question, assign a label from [’Description’, ’Entity’,  ’Expression’, ’Human’, ’Location’, ’Number’]. Return label only without any other text.

  Example1:
  Sentence: "Calculus is a branch of mathematics that deals with rates of change and slopes of curves. It includes differential and integral calculus. The fundamental theorem of calculus links these two concepts."
  Label: Description

  Example2:
  Sentence: "Who wrote the screenplay for the movie where the main character delivers a famous monologue about the American Dream?"
  Label: Human

  Example3:
  Sentence: "The human genome consists of all the genetic information in a human cell. It is composed of approximately 3 billion base pairs."
  Label: Number

  Example4:
  Sentence: "The Magna Carta, signed in 1215, was a landmark document in English history."
  Label: Description

  Example5:
  Sentence: "The director chose to shoot the film in black and white to evoke a sense of nostalgia."
  Label: Description

  Example6:
  Sentence: "Calculus involves the study of limits, derivatives, integrals, and infinite series. It is essential for understanding changes in various quantities."
  Label: Description

  Example7:
  Sentence: "She delivered the line with such conviction it seemed real. The audience was moved."
  Label: Expression

  Example8:
  Sentence: "Which director is known for their innovative use of camera angles in films?"
  Label: Human

  Example9:
  Sentence: "Meryl Streep has performed in many famous plays."
  Label: Location

  Example10:
  Sentence: "He directed the actors to bring out the raw emotion in their performances. The result was powerful."
  Label: Expression

  Example11:
  Sentence: "The screenplay's dialogue was sharp and witty, setting the tone for the entire film."
  Label: Expression

  Example12:
  Sentence: "Newton's laws of motion describe how objects move under the influence of forces."
  Label: Description

  Example13:
  Sentence: "The screenplay features complex dialogue that drives the characters' motivations and relationships."
  Label: Description

  Example14:
  Sentence: "Recent studies show that vitamin C is crucial for the immune system. It helps in fighting infections."
  Label: Description

  Example15:
  Sentence: "William Shakespeare's plays, such as 'Hamlet' and 'Macbeth,' are considered masterpieces of English literature. They explore complex themes like ambition, revenge, and madness."
  Label: Description
}{teal}
\promptbox{SUBJ}{Please perform Subjectivity Classification task. Given the sentence, assign a label from [’subjective’,  ’objective’]. Return label only without any other text.

  Example1:
  Sentence: "The soundtrack was moving."
  Label: subjective

  Example2:
  Sentence: "The soundtrack added an emotional depth."
  Label: subjective

  Example3:
  Sentence: "The pacing started slow. It built tension. The climax felt rushed."
  Label: subjective

  Example4:
  Sentence: "The visual effects were impressive. However, a few scenes felt overdone. Enhanced the world-building but occasionally distracted from the story."
  Label: subjective

  Example5:
  Sentence: "The pacing was uneven. The first half dragged while the second felt rushed."
  Label: subjective

  Example6:
  Sentence: "The visual effects were impressive, though a few scenes felt overdone. They enhanced the world-building but occasionally distracted from the story."
  Label: subjective

  Example7:
  Sentence: "The pacing started slow but built tension." Sentence: "Climax felt rushed." Sentence: "Overall, uneven."
  Label: subjective

  Example8:
  Sentence: "The lead actor brought depth to the role."
  Label: subjective

  Example9:
  Sentence: "The pacing started slow. It built tension. The climax felt rushed."
  Label: subjective

  Example10:
  Sentence: "The lead actress gave a nuanced performance."
  Label: subjective

  Example11:
  Sentence: "The editing was choppy, disrupting the flow. It felt rushed at times."
  Label: subjective

  Example12:
  Sentence: "The screenplay was tightly constructed. The dialogue flowed naturally. Characters spoke authentically, enhancing the plot."
  Label: subjective

  Example13:
  Sentence: "The screenplay was tight. The dialogue flowed naturally. Characters spoke authentically, enhancing the plot. Subtle hints of conflict kept the audience engaged."
  Label: subjective

  Example14:
  Sentence: "The soundtrack added an emotional depth. However, the choice of music was sometimes jarring."
  Label: subjective}{teal}

\section{Simplification Prompt}
\label{sec:sim_prompt}

In this section, we present the best-performing prompts for \ours{} on the simplification tasks. 
The base prompts are adapted from \citep{guo12connecting}, while the examples are generated using our method.

\promptbox{SIMPLIFICATION}{Simplify the text.

  Example1:
  Complex: "The Supreme Court decision declared the law unconstitutional, invalidating it."
  Simple: "The Supreme Court declared the law unconstitutional."

  Example2:
  Complex: "Mount Everest is the highest mountain in the world located in the Himalayas."
  Simple: "Mount Everest is the highest mountain in the Himalayas."

  Example3:
  Complex: "The pizza place offers a variety of toppings including pepperoni and mushrooms."
  Simple: "The pizza place offers pepperoni and mushrooms."

  Example4:
  Complex: "The Eiffel Tower is a famous landmark in Paris, France."
  Simple: "The Eiffel Tower is in Paris, France."

  Example5:
  Complex: "The university offers a range of degree programs in various fields of study."
  Simple: "The university offers degree programs."

  Example6:
  Complex: "The student passed the exam with excellent grades."
  Simple: "The student passed with excellent grades."

  Example7:
  Complex: "The basketball game was won by the team with the highest score at the end of the game."
  Simple: "The team with the highest score won."

  Example8:
  Complex: "The Renaissance was a period of great cultural change and achievement that started in Italy in the 14th century."
  Simple: "The Renaissance started in Italy in the 14th century."

  Example9:
  Complex: "The Industrial Revolution began in the late 18th century and changed manufacturing methods."
  Simple: "The Industrial Revolution changed manufacturing methods in the late 18th century."

  Example10:
  Complex: "The Mona Lisa is a famous painting by Leonardo da Vinci."
  Simple: "The Mona Lisa is a famous painting."

  Example11:
  Complex: "The internet is a global network that connects computers and allows for communication."
  Simple: "The internet connects computers for communication."

  Example12:
  Complex: "The Supreme Court ruled that the law was unconstitutional."
  Simple: "The Supreme Court said the law was unconstitutional."

  Example13:
  Complex: "The Renaissance art focused on humanism and realism, emphasizing individual expression and naturalism."
  Simple: "Renaissance art emphasized individual expression."

  Example14:
  Complex: "The train arrived late due to a track problem."
  Simple: "The train was late due to a track problem."

  Example15:
  Complex: "The internet protocol is a set of rules that allows computers to communicate over the internet."
  Simple: "The internet protocol allows computers to communicate."

}{teal}

\section{Summarization Prompts}
\label{sec:sum_prompt}
In this section, we present the most effective prompt for \ours{} on summarization tasks. 
The base prompts are adapted from \citep{guo12connecting}, and the examples are generated by our method.

\promptbox{SUMMARIZATION}{How would you rephrase that in a few words?

Example1:
Text: ""IBM is an American multinational technology company headquartered in Armonk, New York.""
Summary: ""IBM is headquartered in New York.""

Example2:
Text: ""Apple Inc. is an American multinational technology company headquartered in Cupertino, California.""
Summary: ""Apple Inc. is headquartered in California.""

Example3:
Text: ""Tesla is an American electric vehicle and clean energy company.""
Summary: ""Tesla is an electric vehicle company.""

Example4:
Text: ""The NBA All-Star Game is an annual basketball game featuring the top players from the National Basketball Association.""
Summary: ""The NBA All-Star Game features top NBA players.""

Example5:
Text: ""Global warming is caused by an increase in greenhouse gases, leading to rising temperatures and climate changes.""
Summary: ""Global warming is caused by rising greenhouse gas levels.""

Example6:
Text: ""Tesla is an American electric vehicle and clean energy company founded by Elon Musk, known for its innovative electric cars and energy storage solutions.""
Summary: ""Tesla is an electric vehicle company.""

Example7:
Text: ""The Supreme Court justices are appointed by the President and confirmed by the Senate, serving life terms.""
Summary: ""Supreme Court justices are appointed by the President and confirmed by the Senate.""

Example8:
Text: ""The Supreme Court of the United States is the highest court in the country, responsible for interpreting the Constitution and ensuring federal laws are followed.""
Summary: ""The Supreme Court of the United States interprets the Constitution.""

Example9:
Text: ""The Mona Lisa, a painting by Leonardo da Vinci, is one of the most famous and most visited paintings in the world, currently housed in the Louvre Museum in Paris.""
Summary: ""The Mona Lisa is a famous painting by Leonardo da Vinci.""

Example10:
Text: ""The Louvre Abu Dhabi, opened in 2017, is a museum in Abu Dhabi that focuses on art and culture from around the world.""
Summary: ""The Louvre Abu Dhabi opened in 2017 and focuses on global art and culture.""

Example11:
Text: ""The Louvre Museum in Paris is one of the largest and most visited art museums in the world, with a vast collection of art and artifacts.""
Summary: ""The Louvre Museum in Paris has a large art collection.""

Example12:
Text: ""The American Revolution was a violent conflict between Great Britain and thirteen of its North American colonies from 1775 to 1783.""
Summary: ""The American Revolution lasted from 1775 to 1783.""

Example13:
Text: ""The Renaissance was a period of great cultural and intellectual growth in Europe, spanning the 14th to the 17th century.""
Summary: ""The Renaissance was a period of cultural and intellectual growth in Europe.""

Example14:
Text: ""The Renaissance was a period of great cultural and intellectual growth in Europe, spanning the 14th to the 17th century, marked by a revival of classical learning.""
Summary: ""The Renaissance was a period of cultural and intellectual growth in Europe.""

Example15:
Text: ""The Eiffel Tower is a wrought-iron lattice tower on the Champ de Mars in Paris, France.""
Summary: ""The Eiffel Tower is in Paris, France.""

Example16:
Text: ""The giant panda is a bear species endemic to central China, recognized by its distinctive black and white fur and diet primarily consisting of bamboo.""
Summary: ""The giant panda is a bear species endemic to China.
}{teal}

\section{Pseudocodes}
\label{sec:pseudocode}

We present concise pseudocodes for our method and its Shapley-driven oracle. 
Algorithm~\ref{alg:fast-icl-shapley} orchestrates the full crafting loop: starting from $k$ initial examples, it performs $I$ rounds that evaluate on a small subsample (size $s$) augmented with a replay buffer (size $r$). In each round, \textsc{MonteCarloShapleyWorst} identifies the least helpful example using $P$ random permutations, the improver proposes $m$ candidates, and a conservative \textsc{Replace/Drop/Keep} rule updates the set only when accuracy does not regress. 
Algorithm~\ref{alg:mc-shapley-worst} details the Shapley routine with memoized coalition values $v(S)$ and permutation-averaged marginal contributions, returning the index with the smallest estimated value. 

\section{Usage of LLMs}

We used LLMs to help polish the presentation and writing in this manuscript and to assist in drafting portions of the implementation code; all substantive research decisions and core technical contributions, however, were made by the authors.

\begin{algorithm*}[ht]
\caption{\ours{}}
\label{alg:fast-icl-shapley}
\begin{algorithmic}[1]
\Require
\Statex Dataset $\mathcal{D}$
\Statex Example Proposer $M_{\text{prop}}$
\Statex Prompt Evaluator $M_{\text{eval}}$
\Statex Example Improver $M_{\text{impr}}$
\Statex $k$ (Initial examples) 
\Statex $I$ (Number of craft iterations) 
\Statex $m$ (Number of refine candidates)
\Statex $s$ (Size of subdataset)
\Statex $r$ (Replay Size) 
\Statex $P$ (Number of Shapley Permutations)
\Ensure Crafted example set $E^\star$
\Statex

\State $E \gets \Call{ProposeInitialExamples}{M_{\text{prop}}, k}$ \Comment{$|E|=k$}
\State $R \gets \varnothing$ \Comment{Replay buffer}
\For{$t = 0,1,\dots,I-1$}
    \State $D_t \gets \Call{Subsample}{\mathcal{D}_{\text{infer}}, s}$ 
    \State $\tilde{D}_t \gets D_t \cup R$ \Comment{Union with replay}
    \State $a_{\text{base}} \gets \Call{EvalAcc}{M_{\text{eval}}, E, \tilde{D}_t}$
    \State $i^\star \gets \Call{MonteCarloShapleyWorst}{E, \tilde{D}_t, M_{\text{eval}}, P}$
    \State $E_{\setminus i^\star} \gets E \setminus \{e_{i^\star}\}$
    \State $a_{\text{drop}} \gets \Call{EvalAcc}{M_{\text{eval}}, E_{\setminus i^\star}, \tilde{D}_t}$
    \State $C \gets \Call{GenerateCandidates}{M_{\text{impr}}, E_{\setminus i^\star}, m}$
    \State $(c^{\text{best}}, a_{\text{best}}) \gets \arg\max\limits_{c \in C}\; \Call{EvalAcc}{M_{\text{eval}}, E_{\setminus i^\star} \cup \{c\}, \tilde{D}_t}$
    \Statex \Comment{\textbf{Decision}: \textsc{Replace} vs \textsc{Drop} vs \textsc{Keep}}
    \If{$a_{\text{best}} \ge a_{\text{drop}}$ \textbf{and} $a_{\text{best}} \ge a_{\text{base}}$}
        \State $E \gets E_{\setminus i^\star} \cup \{c^{\text{best}}\}$ \Comment{\textsc{Replace}}
    \ElsIf{$a_{\text{drop}} \ge a_{\text{base}}$ \textbf{and} $|E|>1$}
        \State $E \gets E_{\setminus i^\star}$ \Comment{\textsc{Drop}}
    \Else
        \State $E \gets E$ \Comment{\textsc{Keep}}
    \EndIf
    \State $R \gets R \cup \Call{SampleReplay}{D_t, r}$
\EndFor
\State $E^\star \gets E$
\State \Return $E^\star$
\end{algorithmic}
\end{algorithm*}

\begin{algorithm*}[ht]
\caption{\textsc{MonteCarloShapleyWorst}}
\label{alg:mc-shapley-worst}
\begin{algorithmic}[1]
\Require
\Statex Example set $E=\{e_1,\dots,e_n\}$
\Statex Dataset subset $\tilde{D}$
\Statex Prompt Evaluator $M_{\text{eval}}$
\Statex $P$ (Number of Shapley permutations)
\Ensure Worst index $i^\star$
\Statex

\State $\mathcal{V} \gets \varnothing$ 
\State For each $i \in [n]$, set list $\Delta_i \gets [\;]$
\State Define $v(S) \gets \Call{EvalAcc}{M_{\text{eval}}, \{e_j : j\in S\}, \tilde{D}}$
\State $\mathcal{V}[\varnothing] \gets v(\varnothing)$

\For{$p = 1,2,\dots,P$}
  \State $\pi \gets$ a random permutation of $[n]$
  \State $S \gets \varnothing$;\; $v_{\text{prev}} \gets \mathcal{V}[\varnothing]$
  \For{$j = 1,2,\dots,n$}
    \State $i \gets \pi_j$;\; $S' \gets S \cup \{i\}$
    \If{$S' \notin \mathcal{V}$}
      \State $\mathcal{V}[S'] \gets v(S')$
    \EndIf
    \State $v_{\text{new}} \gets \mathcal{V}[S']$
    \State Append $(v_{\text{new}} - v_{\text{prev}})$ to $\Delta_i$ \Comment{Marginal contribution of $e_i$}
    \State $S \gets S';\; v_{\text{prev}} \gets v_{\text{new}}$
  \EndFor
\EndFor

\For{$i = 1,2,\dots,n$}
  \State $\phi_i \gets \begin{cases}
  \frac{1}{|\Delta_i|}\sum\limits_{d \in \Delta_i} d, & |\Delta_i|>0\\
  0, & \text{otherwise}
  \end{cases}$
\EndFor
\State $i^\star \gets \arg\min\limits_{i \in [n]} \phi_i$
\State \Return $i^\star$
\end{algorithmic}
\end{algorithm*}

\end{document}